\documentclass[sigconf,nonacm]{acmart}

\usepackage{algorithm}
\usepackage{algorithmic}

\AtBeginDocument{%
  }

\AtBeginDocument{%
  \fancypagestyle{standardpagestyle}{%
    \fancyhf{}%
    \fancyfoot[C]{\thepage}%
  }%
  \fancypagestyle{firstpagestyle}{%
    \fancyhf{}%
    \fancyfoot[C]{\thepage}%
  }%
  \pagestyle{standardpagestyle}%
}

\begin{document}

\title{CAS-FD: Contact-Aware Temporal Sampling for Single-View Foul vs. Dive Recognition}

\author{Md. Jahidul Islam}
\email{jahid.prog@gmail.com}
\affiliation{%
  \institution{Premier University}
  \department{Department of Computer Science}
  \city{Chattogram}
  \country{Bangladesh}
}

\author{Mahfujul Alam}
\email{mahfujsakib133@gmail.com}
\affiliation{%
  \institution{Premier University}
  \department{Department of Computer Science}
  \city{Chattogram}
  \country{Bangladesh}
}

\author{Md. Nazmul Islam Seyam}
\email{nazmulseyam6393@gmail.com}
\affiliation{%
  \institution{Premier University}
  \department{Department of Computer Science}
  \city{Chattogram}
  \country{Bangladesh}
}

\author{Md. Tamim Hossain}
\authornote{Corresponding author.}
\email{tamim.hossain@puc.ac.bd}
\orcid{} 
\affiliation{%
  \institution{Premier University}
  \department{Department of Computer Science}
  \city{Chattogram}
  \country{Bangladesh}
}

\renewcommand{\shortauthors}{Islam et al.}

\begin{abstract}
Distinguishing a genuine foul from a simulated dive in football
remains one of the sport's most contested fine-grained recognition
problems, especially when such decisions have to be from a single broadcast
view without multi-view camera angle. Current sports video benchmarks do
not address this setting, and to the best of our knowledge there is no
publicly documented single-view dataset exists for the task. We introduce
a \textbf{balanced 600-clip single-view} Foul/Dive dataset and show that
contact-aware sampling concentrating the model's attention around the
moment of physical contact rather than treating all frames equally yields
substantially improved recognition of this contact-specific problem. The
proposed approach achieves \textbf{86.0\%} accuracy and \textbf{macro-F1
0.860} on the held-out test split, a 12 percentage-point gain over
contact-unaware alternatives that grows further on unseen data. We also
evaluate each pipeline component against human annotations, establishing
where and why the system succeeds and fails. The result is a documented
dataset, a reproducible single-view pipeline, and a grounded evaluation
framework for fine-grained contact-event recognition in broadcast football
footage. The dataset and code are available at
\href{https://github.com/hossain-tamim/contact-aware-dive}{https://github.com/hossain-tamim/contact-aware-dive}.
\end{abstract}



\maketitle

\section{Introduction}
\label{sec:intro}

Whether a player has been fouled or has dived is one of the most
controversial elements of association football, and it regularly
captures the attention of broadcasters, governing bodies, and
fans.  Even with the introduction of the Video Assistant Referee
(VAR), the decision remains subjective, since the visual evidence
available to the referee is limited to a few seconds of footage
that often contains motion blur, partial occlusion, and
fast-panning cameras~\cite{held2024vars}.  Automating, or at least assisting, this
judgement is therefore a worthwhile target for computer-vision
research.

A clip of a foul or a dive is a short, fast, fine-grained action:
the discriminative information is dense in a fraction of a second
around the moment when contact is made (or not)~\cite{weiss2012kinematic}.  A standard
uniform-sampling approach wastes capacity on frames in which
little of interest occurs --- the run-up, the aftermath, the
celebration --- and sometimes misses the contact moment
altogether.

Recent work has shown that several synchronised camera views
enable fine-grained recognition of fouls, such as the
SoccerNet-MVFouls benchmark~\cite{held2024varchallenge} and the
VARS~\cite{held2023vars} system.  Multi-view aggregation, however,
requires the broadcast-production infrastructure of a top-tier
match (live, replay, and tactical cameras), and that
infrastructure is precisely what is \emph{not} available at the
moment a referee, a lower-league official, a broadcaster without
multi-camera production, or a viewer on a mobile device actually
has to make a judgement.  The realistic deployment surface is a
single-view broadcast video.  Existing work in this regime has
focused on action \emph{spotting} (locating events in long
matches~\cite{giancola2018soccernet,deliege2021soccernetv2,cioppa2020context}) or on
the full multi-class action vocabulary, not on the binary
foul-vs-dive distinction at the deep-learning scale.  Earlier
single-view work by Al-Theiabat and Aljarrah~\cite{altheiabat2018dive}
addresses the dive-vs-foul question on $25$ manually pre-selected
tackle clips with a classical pipeline (manual player
localisation, optical-flow tracking, NBTree classifier on
hand-engineered velocity features), reporting $84\%$ accuracy
under 10-fold cross-validation but with all four mis-classifications
being fouls predicted as dives.  Their dataset is not publicly
released and the pipeline requires per-clip human intervention,
so a deep-learning, automated, larger-scale single-view
Foul/Dive corpus and pipeline does not yet exist~\cite{rao2020fouldive,bai2021skeleton,naik2022dive}.

\paragraph*{Scope and intent.}
This paper is an \emph{engineering} contribution rather than an
architectural one.  We do not propose a new backbone, loss, or
trainable component; the components of our pipeline (YOLO
detection, signal smoothing, max-of-combined-score, priority
centroid) are individually standard.  What we add is (i) a
documented dataset of a size and form not previously released
for the single-view Foul/Dive task, (ii) a particular composition
of those standard components for that task, with no manual
per-clip intervention, and (iii) a qualitative evaluation that
compares the front-end blocks against human-annotated ground
truth.  The contributions are:
\begin{itemize}
  \item A small-scale ($600$-clip), balanced single-view
        Foul/Dive dataset complementing the multi-view
        SoccerNet-MVFouls benchmark in the deployment regime
        where replay angles are unavailable, and approximately
        $24\times$ larger than the only previous single-view
        Foul/Dive corpus~\cite{altheiabat2018dive}, which is not
        publicly released.  We do not claim our dataset is
        large; we claim it is the first such corpus to be
        publicly documented at this scale.
  \item A YOLO-driven contact-aware temporal sampler and
        priority-weighted spatial crop, fully specified
        (Sec.~\ref{sec:method}) and ablated against three
        content-blind baselines on both validation and held-out
        test splits, with a three-seed reproducibility check on
        the head of the leaderboard.  Unlike the optical-flow
        approach of Al-Theiabat and
        Aljarrah~\cite{altheiabat2018dive}, the pipeline
        operates end-to-end on raw broadcast clips with no
        manual selection of the falling player.
  \item A per-component evaluation: predicted contact frames
        compared against $20$ human-annotated contact frames
        ($10$ Foul, $10$ Dive), with the model's asymmetric
        $16$-frame window covering the human-annotated frame in
        $19$ of $20$ cases; and predicted crops compared
        against the averaged box of three independent student
        annotations on the same $20$ clips, with mean IoU
        $0.369 \pm 0.207$ against an inter-annotator ceiling of
        $0.965 \pm 0.062$.
\end{itemize}

The remainder of the paper is organised as follows.
Section~\ref{sec:related} reviews the relevant literature.
Section~\ref{sec:method} describes the dataset, the four samplers,
and the training framework.  Section~\ref{sec:experiments} details
the experimental protocol.  Section~\ref{sec:comparison} reports
the comparative results.  Section~\ref{sec:discussion} discusses
implications and limitations, and Section~\ref{sec:conclusion}
concludes.

\section{Related Work}
\label{sec:related}

\subsection{Video Action Recognition}
Modern video action recognition is dominated by transformer
architectures that lift the Vision Transformer~\cite{dosovitskiy2021vit}
to the spatio-temporal domain, including
TimeSformer~\cite{bertasius2021timesformer}, Video
Swin~\cite{liu2022videoswin}, ViViT~\cite{arnab2021vivit}, and
MViT~\cite{fan2021mvit}.  Prior to the transformer era, two-stream
networks~\cite{simonyan2014twostream} processed RGB frames and
optical flow in parallel, and 3D convolutional architectures such
as C3D~\cite{tran2015c3d} and I3D~\cite{carreira2017quovadis}
established the inflation paradigm for transferring 2D ImageNet
weights to volumetric video features.  Self-supervised pre-training
has proved particularly effective in the data-scarce regime:
the masked autoencoder paradigm~\cite{he2022mae} was extended to
video by VideoMAE~\cite{tong2022videomae}, which masks a high
fraction of $3$D tubelets and reconstructs them, learning
representations that transfer well to small downstream datasets;
VideoMAE V2~\cite{wang2023videomaev2} subsequently scales this
approach with dual masking.  Subsequent extensions such as
EVEREST~\cite{hwang2024everest} have shown that masked video
autoencoders can be made markedly more efficient by discarding
uninformative spatio-temporal cubes during pre-training, but our
small-scale fine-tuning regime does not require this optimisation.
Because our task has only $400$ training clips, we adopt
VideoMAE-Base pre-trained on Kinetics-$400$~\cite{kay2017kinetics} as the backbone.

\subsection{Temporal Sampling}
Uniform sampling, used as the default in TSN~\cite{wang2016tsn} and
SlowFast~\cite{feichtenhofer2019slowfast}, is simple and
deterministic but is known to under-sample short fine-grained
actions.  Adaptive samplers such as
SCSampler~\cite{korbar2019scsampler} and
AdaFrame~\cite{wu2019adaframe} use a lightweight network to score
frames or to early-exit, and the family extends to per-frame
resolution selection~\cite{meng2020arnet} and efficient two-stage
schemes such as ECO~\cite{zolfaghari2018eco}; all of these are
trained jointly with the action classifier and require a
non-trivial annotation budget.  Our contact-aware sampler is
closer in spirit to the content-based heuristics used in
event-spotting~\cite{giancola2018soccernet} and event-anchored
sampling for contact-sport recognition~\cite{hussain2023eventanchored},
where domain-specific signals direct the temporal window without
requiring extra labels.

\subsection{Multi-View Foul Recognition}
Held et al.\ introduced foul recognition as a dedicated
computer-vision task with the VARS system~\cite{held2023vars} and
the SoccerNet-MVFouls dataset~\cite{held2024varchallenge}, which
collects approximately $3{,}900$ foul actions captured by two to
four synchronised camera views and annotates each action with
ten properties.  The action-type label space includes eight
fine-grained classes --- Standing Tackle, Tackle, Holding,
Pushing, Challenge, Dive, High Leg, and Elbowing --- so a
``Dive'' annotation does exist in the multi-view setting; what
is absent is a single-view binary release of the same examples.
Subsequent SoccerNet challenges~\cite{cioppa2024soccernet} have
built on this benchmark, and the most recent multi-view VARS
reports correct foul-type classification on around half of
incidents and correct sanction prediction on roughly $46\%$ of
cases, all under the assumption that multiple synchronised
camera angles are available at inference time~\cite{held2024vars}.
The defining methodological contribution of this line of work is
multi-view aggregation; the setting we consider is strictly more
constrained.

\subsection{Single-View Foul/Dive Recognition}
\label{sec:related-singleview}
To our knowledge, the only previously published  
single-view
Foul/Dive system is the computer-vision pipeline of Al-Theiabat
and Aljarrah~\cite{altheiabat2018dive}.  Their system works on
$25$ manually curated tackle clips ($12$ Foul, $13$ Dive) and
proceeds in five stages: manual mouse-bounded selection of the
falling player, Shi--Tomasi corner-point extraction
on that player, optical-flow tracking with the
Kanade--Lucas--Tomasi method augmented by pyramid levels and
forward--backward error filtering, computation of two
hand-engineered features (average and maximum vertical
velocity of the player centre of mass), and classification
with a Weka NBTree on those two features under $10$-fold
cross-validation, reaching $84\%$ accuracy.  Their reported
confusion matrix is asymmetric: all $13$ Dive samples are
classified correctly, but $4$ of $12$ Foul samples are
mis-classified as Dive --- the same Foul-recall deficit that
content-blind sampling exhibits on contact-driven actions
(Sec.~\ref{sec:perclass}).  Related single-view work has
explored two-stream deep classifiers on broadcast
clips~\cite{rao2020fouldive}, skeleton-based dive detection
using OpenPose features and bidirectional
LSTMs~\cite{bai2021skeleton,cao2021openpose}, transformer-based
dive detection on proprietary clip sets~\cite{naik2022dive},
and tracking-data-driven simulation
classifiers~\cite{bransen2019tracab}; complementary
biomechanical studies have characterised the kinematic
differences between genuine and simulated
falls~\cite{weiss2012kinematic}.  Our work differs from these
prior single-view efforts in four respects.
First, the dataset is approximately $24\times$ larger than~\cite{altheiabat2018dive} and is
documented in a single-view setting where their dataset is not
publicly released.  Second, the pipeline is end-to-end automated:
no per-clip mouse interaction is required, because YOLO
detection drives both temporal and spatial pre-processing.
Third, the classifier is a modern self-supervised video
backbone (VideoMAE-Base, fine-tuned), rather than a two-feature
NBTree.  Fourth, beyond end-to-end accuracy, we evaluate the
two front-end blocks against human annotation.  These
differences are scale and methodology rather than novelty in
problem framing; \cite{altheiabat2018dive} was the first to
state the problem in computer-vision terms.

\subsection{Single-View Broadcast Football Video}
A separate body of work targets single-view broadcast footage
but focuses on different tasks.  SoccerNet~\cite{giancola2018soccernet}
and its successors~\cite{deliege2021soccernetv2} treat the
problem as action spotting; other single-view work addresses
ball detection, player localisation, event captioning,
head-impact recognition~\cite{deepimpact2022}, and rule-specific
decisions such as offside calls from a single broadcast
camera~\cite{desai2024offside}.  Efficient video action
recognition tailored to football~\cite{hurault2023evarf} has
further confirmed the suitability of masked-autoencoder
backbones for short football action clips.  A complementary
line of recent work scales soccer video understanding to a
corpus of $1{,}988$ full matches with a visual-language
foundation model~\cite{rao2024unisoccer}; that release is again
designed for general-purpose tasks (event classification,
commentary generation, multi-view foul recognition) rather than
a single-view binary Foul-vs-Dive split.

\subsection{Detection-Driven Pre-processing}
The use of an off-the-shelf detector to drive video pre-processing
is common in action localisation and in fine-grained sports
analytics.  Our use of YOLO~\cite{redmon2016yolo}, in the
YOLOv8 incarnation~\cite{jocher2023yolov8}, as a temporal-
and-spatial guide for VideoMAE follows this lineage; the detector
affects only the input pipeline and does not require modifying
the backbone.  Related work on player and ball detection in
broadcast football~\cite{vats2021football,vandeghen2022kalman}
demonstrates the practical reliability of such detectors as a
front-end to downstream recognition.

\section{Method}
\label{sec:method}

\subsection{Problem Statement}
We address Foul-vs-Dive recognition in the \emph{single-view}
setting.  Given a video clip
$V \in \mathbb{R}^{F \times H \times W \times 3}$, where $F$ is
the number of frames and $H{\times}W$ is its resolution, we want
to predict a binary label
$y \in \{\textsc{Dive}, \textsc{Foul}\}$.  In our setting
$H{=}720$, $W{=}1280$, the frame rate is $25$\,fps, and the clip
length is fixed at $5$\,s, giving $F{\approx}125$.  The
downstream model requires $T{=}16$ frames at $224{\times}224$.

\subsection{Backbone}
We take pre-trained weights from
VideoMAE-Base~\cite{tong2022videomae} (the ViT-Base
configuration with $16{\times}16$ patches and $224{\times}224$
input), which is
trained on Kinetics-$400$~\cite{kay2017kinetics}.  It has $86{,}228{,}738$ parameters,
processes $T{=}16$ frames with a tubelet size of $2$, and
produces a $768$-dimensional feature.  We swap out the
classifier head for a linear layer with two outputs and
fine-tune the entire backbone with a layer-wise learning-rate
decay of $0.75$~\cite{bao2022beit}.

\subsection{Temporal Sampling Strategies}
\label{sec:samplers}

Four samplers are compared, each returning $T{=}16$ frame
indices:
\begin{itemize}
  \item \textbf{Uniform.} Evenly spaced indices in $[0,F-1]$.
  \item \textbf{Random.} Indices drawn uniformly without
        replacement from $[0,F-1]$ at each epoch.
  \item \textbf{Random window.} A random contiguous sub-clip of
        length $T'>T$ is drawn, and $T$ uniformly-spaced indices
        are taken inside it.
  \item \textbf{Contact-aware (proposed).} Detailed below.
\end{itemize}

The contact-aware sampler runs a YOLO ball-and-player tracker on
each frame and computes ten per-frame signals
(Table~\ref{tab:signals}), all min--max normalised within the
clip.  A combined score
\begin{equation}
  s_f \;=\; \sum_{k=1}^{10} w_k\,\widehat{x}^{(k)}_f
  \label{eq:score}
\end{equation}
is calculated with the weights of Table~\ref{tab:signals}; the
hat indicates per-clip normalisation.  The series $s_f$ is
filtered with a moving average of width $w{=}3$ frames so that a
single detection spike cannot dominate the estimate of contact.

\begin{table}[!t]
\caption{The ten per-frame signals comprising Eq.~\ref{eq:score}
and their weights.  The first four (dominant) signals carry
$0.67$ of the total weight; the remaining six carry $0.33$.}
\label{tab:signals}
\centering
\setlength{\tabcolsep}{4pt}
\renewcommand{\arraystretch}{1.05}
\begin{tabular}{clc}
\toprule
\textbf{$k$} & \textbf{Signal $\widehat{x}^{(k)}_f$} & \textbf{$w_k$} \\
\midrule
$1$ & $\widehat{v}_f$\;: holder--opponent approach velocity              & $0.22$ \\
$2$ & $\widehat{\ell}_f$\;: leg-region IoU between two closest players   & $0.18$ \\
$3$ & $\widehat{\phi}_f$\;: fall onset (drop in height/width ratio)      & $0.15$ \\
$4$ & $|\Delta\widehat{\phi}_f|$\;: ratio velocity                       & $0.12$ \\
\midrule
$5$ & inverse holder--opponent distance (proximity)                      & $0.10$ \\
$6$ & body-IoU between holder and nearest opponent                       & $0.08$ \\
$7$ & holder-bbox area frame-to-frame change (size drop)                 & $0.07$ \\
$8$ & ball-bbox displacement frame-to-frame                              & $0.05$ \\
$9$ & inverse referee--players distance (referee approach)               & $0.02$ \\
$10$ & frame-level mean pixel motion                                     & $0.01$ \\
\midrule
& \textbf{Total}                                                         & $1.00$ \\
\bottomrule
\end{tabular}
\end{table}

\paragraph*{Approach-window guard.}
The estimated contact instant is taken from the maximum of
$\tilde{s}$ \emph{subject to an approach-window guard}: a
candidate frame $f^\star$ is accepted only if at least three
frames of monotonic inter-player closing motion can be observed
within the previous $6$-frame guard window; otherwise the
next-best candidate is tried.  If no candidate meets the guard,
the system falls back to the raw $\arg\max$.

\paragraph*{Frame layout.}
The $T{=}16$ frames are laid out \emph{asymmetrically} around
$f^\star$ --- nine before the contact frame, the contact frame
itself, and six after, at stride $\tau{=}3$, covering
$\approx 1.8$\,s.  The asymmetric layout was chosen over a
symmetric $8{+}8$ layout because the approach phase
(closing velocity, intent indicators, body positioning) is more
discriminative for the foul/dive distinction than the aftermath
(the fall, the referee reaction), which is visually similar
across the two classes~\cite{weiss2012kinematic}.

\subsection{Adaptive Crop}
\label{sec:adaptive-crop}
The same detector drives a priority crop of size $224{\times}224$.
For the $T$ selected frames a per-frame weight is computed as the
product of a Gaussian decay around $f^\star$ and the combined
signal $\tilde{s}_f$, with anchor priorities
$w_{\text{ball}}{=}3.0$, $w_{\text{holder}}{=}2.5$,
$w_{\text{opp}}{=}2.0$, $w_{\text{fall}}{=}1.8$,
$w_{\text{leg}}{=}1.5$.  The weighted centroid $c$ specifies the
centre of a square crop whose side is the tightest box covering
the top-$60\%$ weighted evidence, clamped to $[180, 380]$ pixels
at decode resolution ($360{\times}640$, scaling to
$[360, 760]$ pixels at native $1280{\times}720$).  The square
crop is then resized to $256{\times}256$ and centre-cropped to
$224{\times}224$, matching the ImageNet/Kinetics pre-processing
recipe used during VideoMAE pre-training.  Referees are excluded
from anchor selection using a jersey-colour KMeans team
assigner.  If all anchors are absent from a clip ($<2\%$ of
cases), the crop falls back to the frame centre.

\setlength{\textfloatsep}{4pt}
\begin{algorithm}[!t]
\caption{Contact-aware temporal sampling and priority crop.}
\label{alg:contact}
\begin{algorithmic}[1]
\REQUIRE clip frames $V \in \mathbb{R}^{F\times H\times W\times 3}$, target $T$
\STATE $\{B_f\}_{f=0}^{F-1} \leftarrow \mathrm{YOLO}(V)$
       \hfill\emph{// per-frame detections}
\STATE compute $\widehat{x}^{(k)}_f$ for $k = 1,\dots,10$ from $B_f$
\STATE compute $s_f$ via Eq.~\ref{eq:score}
\STATE $\tilde{s} \leftarrow \mathrm{MovingAvg}(s, w{=}3)$
\STATE $f^\star \leftarrow$ first element of $\mathrm{argsort}(-\tilde{s})$
       satisfying approach-window guard
\STATE $\mathcal{I} \leftarrow \{f^\star{-}9\tau,\dots,f^\star,\dots,f^\star{+}6\tau\}$
\STATE $c \leftarrow$ Gaussian-weighted priority centre over $\mathcal{I}$
\STATE $\mathcal{C} \leftarrow$ square crop on $c$, resize-256 then
       centre-crop-224
\RETURN $\bigl\{\, V[f][\mathcal{C}] \,\bigr\}_{f \in \mathcal{I}}$
\end{algorithmic}
\end{algorithm}

\begin{figure*}[!t]
\centering
\includegraphics[width=0.95\textwidth]{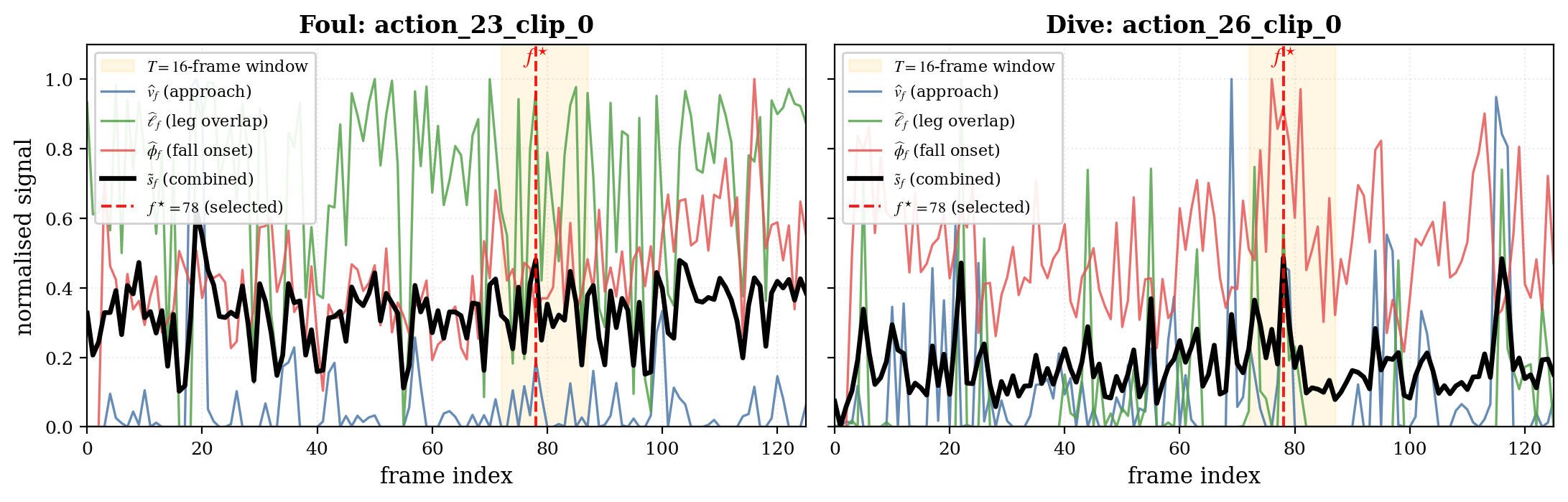}
\caption{Per-frame saliency signals on a representative Foul
(left) and Dive (right) clip.  Plotted: $\widehat{v}_f$,
$\widehat{\ell}_f$, $|\Delta\widehat{\phi}_f|$, and the smoothed
combined score $\tilde{s}_f$ (bold).  Shaded region: the
$T{=}16$-frame window centred on $f^\star$.}
\label{fig:signal-trace}

\end{figure*}

\subsection{Why Detection-Driven?}
\label{sec:crop-design-choice}

An important question is whether the detection-driven adaptive crop is necessary, or if a fixed centre crop is sufficient.  Both
branches share the contact-aware sampler, the same $T{=}16$
frames, and the same resize-$256$/centre-crop-$224$ tail; only
the centroid differs.  The centre crop takes a fixed
$760{\times}760$-pixel square at $(W/2, H/2)$.

\begin{figure}[!t] 
\centering
\includegraphics[width=\columnwidth]{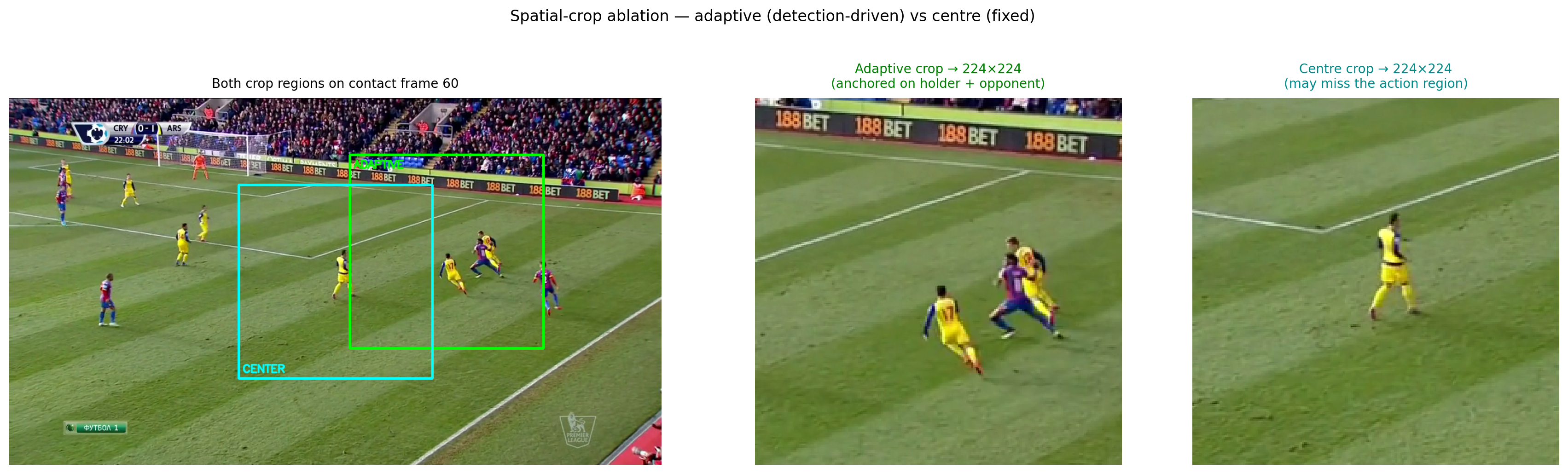}
\caption{Adaptive vs.\ fixed-centre crop on a representative
contact frame.  Left: both regions on the full frame (green:
adaptive, cyan: centre).  Middle/right: the resulting
$224{\times}224$ crops.}
\label{fig:crop-vs-center}
\end{figure}

Figure~\ref{fig:crop-vs-center} shows the comparison.  Broadcast
frames are composed for narrative rather than for
classification, so the action is typically displaced from the
geometric centre.  On this clip the centre crop captures grass
and one off-action player while the foul-causing pair is clipped
at the frame edge; the adaptive crop places those actors at the
centre of the $224{\times}224$ input.  The \emph{quality} of the
adaptive crop is evaluated separately against author annotations
in Sec.~\ref{sec:iou-eval}.

\subsection{Training Framework}
The pipeline is built around a hierarchical YAML configuration:
a single base configuration file contains shared defaults, with
the four sampler-specific files only overriding the
sampling-mode setting.  Each run produces a metadata record, a
resolved configuration, the per-epoch JSON log, the best
checkpoint, and a run summary file, so the four modes are
reproducible in one command per mode.

\begin{figure*}[!t]
\centering
\includegraphics[width=0.95\textwidth]{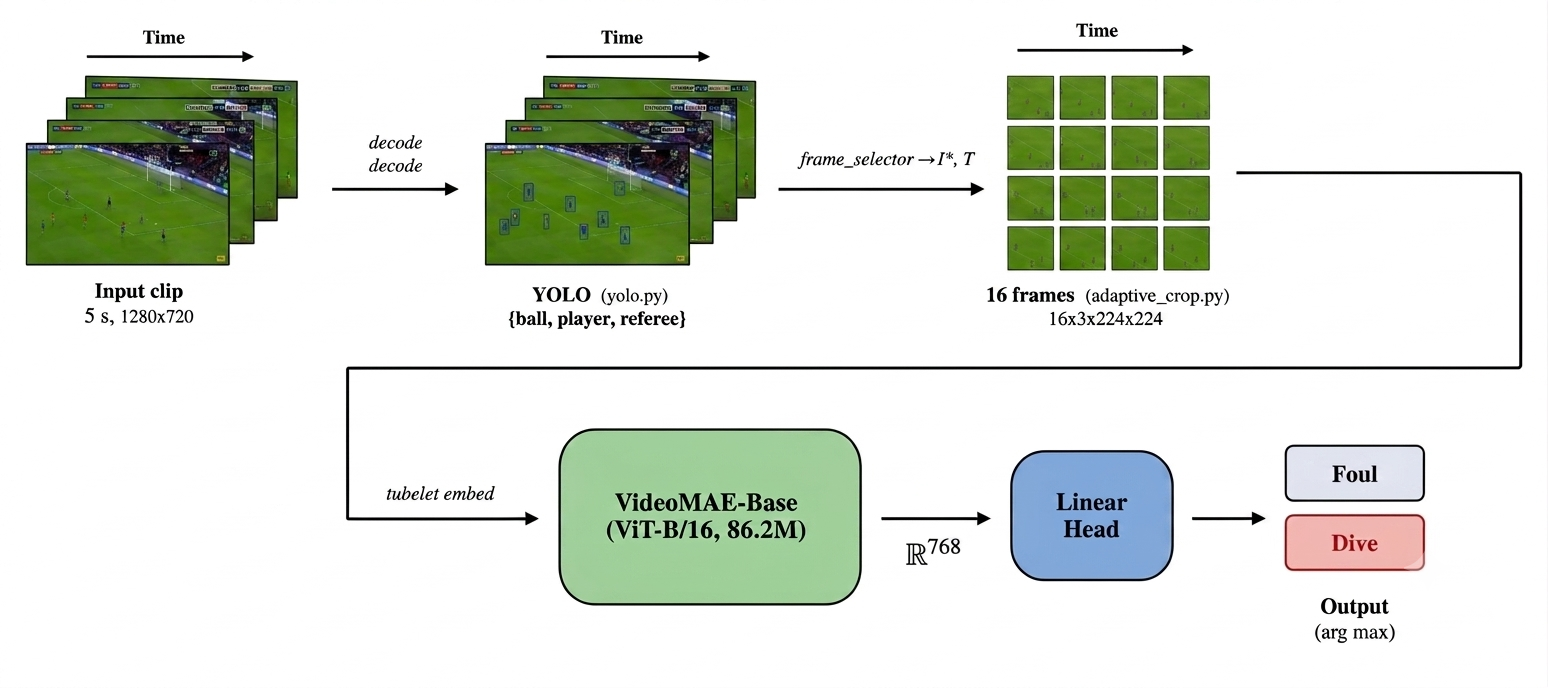}
\caption{Overview of the contact-aware pipeline.
}
\label{fig:pipeline}
\end{figure*}
\section{Experimental Setup}
\label{sec:experiments}

\subsection{Dataset}
\label{sec:dataset}

\paragraph*{Composition.}
We build a balanced single-view Foul/Dive dataset containing
$600$ broadcast football clips, each $5$\,s long at
$1280{\times}720$ and $25$\,fps, split $400$/$100$/$100$ for
train/val/test with class balance enforced in every split.
Files are stored in separate class directories (one for foul
clips and one for dive clips); no frame-level annotations are
required at training time.  The same numeric clip identifier may appear
under both class directories and refers to two different clips
disambiguated by the parent folder.

\paragraph*{Sources.}
Foul clips were extracted from the SoccerNet-v2 spotting
annotations~\cite{deliege2021soccernetv2}, using the labelled
foul timestamps to obtain $5$\,s broadcast segments.  Dive clips
required manual curation because no public single-view Dive
corpus exists.  Candidate dive moments were identified using
the catalogue of curated dive incidents published by the
\emph{Stop Diving} initiative~\cite{stopdiving}, which lists
the broadcasters, matches, and approximate timestamps of
notable documented dives; the corresponding full-broadcast
footage was then trimmed to $5$-second windows of the live feed
around each referenced timestamp, with the contact instant
manually verified by the authors.  Using \emph{Stop Diving}
only as a source of timestamps (not a source of frames) keeps
the dive distribution consistent with the broadcast-feed visual
conditions of the SoccerNet foul clips.

\paragraph*{Scale framing.}
This is a \emph{small-scale, proof-of-concept} dataset and we
make no other claims.  It is two orders of magnitude smaller
than SoccerNet-MVFouls, with a total of $50$\,minutes of
footage.  It is, however, approximately $24\times$ larger than
the $25$-clip dataset of Al-Theiabat and
Aljarrah~\cite{altheiabat2018dive}, which to our knowledge is
the only previous single-view Foul/Dive corpus reported in the
computer-vision literature, and whose clips are not publicly
released.  We aim to enable the first large-enough single-view
Foul/Dive study to support modern fine-tuning protocols.

\begin{figure}[!t]
\centering
\includegraphics[width=\columnwidth]{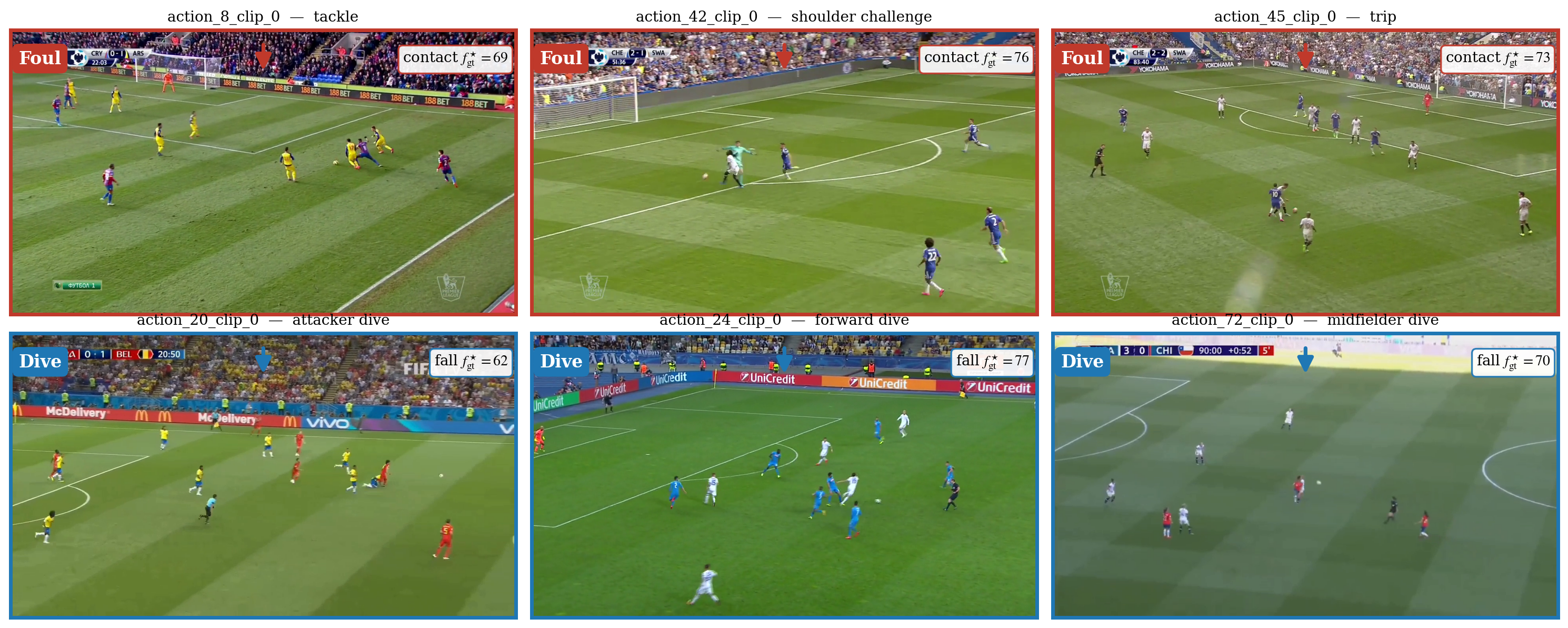}
\caption{Representative single-view broadcast clips. }
\label{fig:dataset}
\end{figure}
\subsection{Implementation Details}
\label{sec:implementation}
We use the VideoMAE-Base model pre-trained on Kinetics-$400$~\cite{kay2017kinetics}
with $T{=}16$, tubelet size $2$, and input resolution
$224{\times}224$.  AdamW~\cite{loshchilov2019adamw} with peak learning rate
$5{\times}10^{-5}$, minimum $10^{-6}$, two-epoch linear warm-up
from $10^{-5}$, and a cosine schedule over $30$ epochs.
Layer-wise LR decay $0.75$~\cite{bao2022beit}, weight decay $0.1$, drop-path $0.2$~\cite{huang2016stochasticdepth}.
Augmentation: RandAugment~\cite{cubuk2020randaugment} (magnitude $7$, $4$ operations per
image, magnitude standard deviation $0.5$, increasing severity),
Mixup~\cite{zhang2018mixup} with $\alpha{=}0.4$, and label smoothing $0.1$~\cite{szegedy2016labelsmoothing}.  Batch
size $4$.  Early stopping is triggered if no validation
improvement is seen after $4$ consecutive epochs.  Video
decoding uses Decord~\cite{decord2019} for efficient
random-access frame access.

\paragraph*{Why $T{=}16$.}
A $16$-frame window at stride $\tau{=}3$ covers
$\approx 1.8$\,s, matching the timescale over which a foul/dive
sequence unfolds: approach ($\approx 0.5$\,s), contact
($\approx 0.04$\,s), and reaction ($\approx 0.5$\,s).  At
$T{=}8$ the window shrinks to $\approx 0.9$\,s, truncating the
approach phase that the contact-aware sampler is specifically
designed to capture (Sec.~\ref{sec:samplers}).  At $T{=}32$ the
window expands to $\approx 3.6$\,s, exceeding the
discriminative interval and effectively reverting toward
uniform-style sampling.  $T{=}16$ is also the configuration in
which VideoMAE-Base is pre-trained on Kinetics-$400$, so this
choice preserves the temporal patch grid and avoids
reinitialising the temporal positional embedding.  Higher
values of $T$ were not ablated under our compute budget.

All four samplers are trained under seed $42$ as the primary
experimental protocol.  For the two best-performing samplers
(contact-aware and uniform) we additionally re-train under
seeds $7$ and $123$ to verify reproducibility of the
seed-$42$ result; these are not separate experimental
conditions but a seed-stability check (Sec.~\ref{sec:repro}).


\subsection{Hardware}
All experiments, including the primary seed-$42$ runs and the
reproducibility runs under seeds $7$ and $123$
(Sec.~\ref{sec:repro}), are performed on a single NVIDIA T$4$
GPU ($16$,GB) on Google Colab Pro with $25$,GB of system RAM.
The training pipeline remains consistent across all runs, with
the data-loader worker count set to zero (single-process
loading) to ensure stable execution within the Colab
environment.

\subsection{Evaluation Protocol}
\label{sec:eval-protocol}
The validation split is used exclusively for model selection
(per-run early stopping and best-epoch checkpointing).  The
$100$-clip test split is held out from selection and is touched
only once per sampler at the end of training, when each run's
best validation checkpoint is evaluated on it to obtain
per-class precision, recall, and F$1$.  Validation accuracies
(Sec.~\ref{sec:comparison}, Table~\ref{tab:results}) appear in
the paper for training-dynamics figures and for the multi-seed
reproducibility check (Sec.~\ref{sec:repro}); the test-set
per-class numbers (Table~\ref{tab:f1-test}) are the headline
final result.  We report Top-$1$ accuracy as the primary metric
and macro-F$1$ as the class-balanced complement.

\subsection{Per-Component Qualitative Evaluation}
\label{sec:per-component-protocol}
End-to-end accuracy alone cannot reveal whether the two
front-end components --- the frame selector and the adaptive
crop --- are behaving as designed.  We therefore add two
qualitative comparisons against human-annotated ground truth.

\paragraph*{Frame-selector annotation.}
$20$ clips were drawn from the dataset ($10$ Foul, $10$ Dive).
The annotator (the first author) watched each clip at the
native $1280{\times}720$ resolution and scrubbed through it
frame-by-frame in a video player to mark the contact instant.
An important semantic distinction was made: in football,
players make incidental physical contact constantly --- run-up
jostling, shoulder-to-shoulder play, leg brushes --- and not
all contact is foul-causing~\cite{ifab2023laws}.  The annotator therefore marked
the contact event \emph{most directly causing the foul outcome}
(in Foul clips) or the onset of the simulated fall (in Dive
clips), ignoring earlier incidental contacts.  We compare this
human-marked frame to the contact frame $f^\star_{\text{pred}}$
returned by the \emph{frame selector}, which has no such
semantic notion and simply maximises the smoothed saliency
$\tilde{s}_f$.

\paragraph*{Adaptive-crop annotation.}
For the same $20$ clips, three of the authors (the three
student authors) independently drew a square axis-aligned
bounding box around the offence region on the human-marked
contact frame at native $1280{\times}720$ resolution,
excluding referees, using the Pixlab Annotate web
tool~(\url{https://annotate.pixlab.io/}).  Each annotator
submitted their annotations without consulting the others.
For each clip, the element-wise mean of the three boxes is
taken as the consensus reference, and the model's adaptive
crop $\mathcal{C}_{\text{pred}}$ is compared against it via
intersection-over-union (Eq.~\ref{eq:iou}).  Per-annotator and
inter-annotator IoUs are also computed and reported below as
context.

\paragraph*{Honest framing.}
The frame-selector evaluation was conducted by one author; the adaptive-crop evaluation was conducted by
three of the authors independently to give an inter-annotator
ceiling and a consensus target.  Both evaluations are at a focused 
small-scale level, and we acknowledge the in-team annotation as a 
limitation in Sec.~\ref{sec:limitations}.  The purpose of
these evaluations is diagnostic --- to show what each of the
two front-end blocks does well and where each one fails, in
concrete cases the reader can inspect --- not to establish a
benchmark or a statistically certified claim.

\section{Results}
\label{sec:comparison}

\subsection{End-to-End Comparison}

Table~\ref{tab:results} reports the validation-set ablation
under seed $42$.  The contact-aware sampler achieves the
highest best-epoch accuracy, and its final-epoch accuracy
matches its best, indicating stable convergence; the three
baselines exhibit volatile trajectories whose final accuracy
falls below their best epoch.

\begin{table}[!t]
\caption{Validation-set ablation ($100$ clips, $50$ Foul /
$50$ Dive), seed $42$, identical hyper-parameters across runs.
Used for model selection (early stopping and best-epoch
checkpointing).  Test-set numbers --- with per-class
precision, recall, and F$1$ for all four samplers --- are
reported separately in Table~\ref{tab:f1-test}.}
\label{tab:results}
\centering
\setlength{\tabcolsep}{3pt}
\begin{tabular}{lccccc}
\toprule
\textbf{Sampler} & \textbf{Best} & \textbf{Best} & \textbf{Best} & \textbf{Final} & \textbf{Ep.}\\
                 & \textbf{Acc} & \textbf{Ep.} & \textbf{Loss} & \textbf{Acc}   & \\
\midrule
Cont.-Aware (prop.) & \textbf{95\%} & 6 & \textbf{0.21} & \textbf{95\%} & 10 \\
Uniform             & 88\%          & 7 & 0.40          & 86\%          & 11 \\
Rand.\ Window       & 87\%          & 2 & 0.37          & 80\%          & 6  \\
Random              & 86\%          & 5 & 0.36          & 84\%          & 9  \\
\bottomrule
\end{tabular}
\end{table}

\begin{figure}[!t]
\centering
\includegraphics[width=\columnwidth]{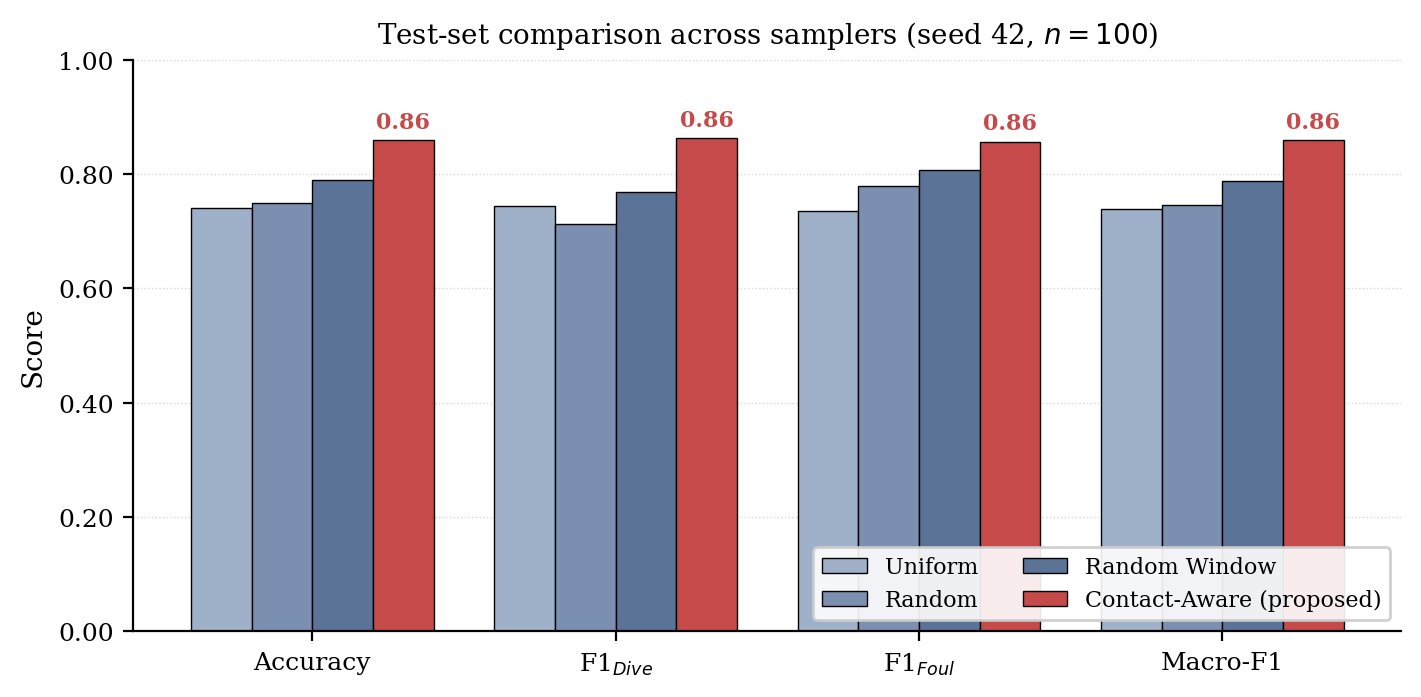}
\caption{Test-set accuracy and per-class F$1$ across the four
samplers (seed $42$, $n{=}100$).}
\label{fig:bar-test}
\end{figure}

Table~\ref{tab:f1-test} reports the test-set evaluation, the
headline final result.  Test predictions for all four samplers
were generated in a single end-to-end inference pass on each
run's best validation checkpoint, allowing direct computation
of per-class precision, recall, and F$1$.
Figure~\ref{fig:bar-test} visualises the comparison.  Two
observations are worth highlighting.  First, the
proposal-vs-baseline gap \emph{widens} on the held-out test
data: the contact-aware sampler reaches $86.0\%$ accuracy and
$0.860$ macro-F$1$, against $74.0\%$ and $0.740$ for uniform, a
$12$\,pp gap (cf.\ $7$\,pp on validation).  Second, the
val-to-test generalisation gap is smallest for the
contact-aware sampler ($9$\,pp) and random-window ($8$\,pp),
and largest for uniform ($14$\,pp), suggesting that
content-driven sampling not only improves accuracy but reduces
over-fitting to the model-selection split.

\begin{table*}[!t]
\caption{Test-set per-class precision, recall, and F$1$ across
the four samplers, seed $42$, on the held-out $100$-clip test
split ($50$ Dive / $50$ Foul).  Best macro-F$1$ in bold.  These
are the final results; the validation accuracies in
Table~\ref{tab:results} were used only for model selection.}
\label{tab:f1-test}
\centering
\setlength{\tabcolsep}{5pt}
\renewcommand{\arraystretch}{1.1}
\begin{tabular}{lcccccccc}
\toprule
\textbf{Sampler} & \textbf{Acc}
  & \textbf{F1$_{\text{Dive}}$} & \textbf{P$_{\text{Dive}}$} & \textbf{R$_{\text{Dive}}$}
  & \textbf{F1$_{\text{Foul}}$} & \textbf{P$_{\text{Foul}}$} & \textbf{R$_{\text{Foul}}$}
  & \textbf{Macro-F$1$} \\
\midrule
Uniform                       & $74.0\%$ & $0.745$ & $0.731$ & $0.760$ & $0.735$ & $0.750$ & $0.720$ & $0.740$ \\
Random                        & $75.0\%$ & $0.713$ & $0.838$ & $0.620$ & $0.779$ & $0.698$ & $0.880$ & $0.746$ \\
Random Window                 & $79.0\%$ & $0.769$ & $0.854$ & $0.700$ & $0.807$ & $0.746$ & $0.880$ & $0.788$ \\
Contact-Aware \emph{(prop.)}  & $\mathbf{86.0\%}$ & $\mathbf{0.863}$ & $0.846$ & $0.880$ & $\mathbf{0.857}$ & $0.875$ & $0.840$ & $\mathbf{0.860}$ \\
\bottomrule
\end{tabular}
\end{table*}

\subsection{Frame Selector vs.\ Human Annotation}
\label{sec:te-eval}

\paragraph*{Annotation protocol.}
We drew $20$ clips from the dataset ($10$ Foul, $10$ Dive) and
one author hand-annotated the contact instant on each one.  An
important distinction was made during annotation: in football,
players make incidental physical contact constantly ---
shoulder-to-shoulder bumps, run-up jostling, leg-on-leg in
normal play --- and not all contact is foul-causing~\cite{ifab2023laws}.  The
human annotator therefore marked the \emph{contact event most
directly causing the foul or dive outcome}, ignoring earlier
incidental contacts.  The contact-aware sampler, by contrast,
has no semantic notion of ``the foul-causing contact'' --- the
\emph{frame selector} simply maximises $\tilde{s}_f$ from
Eq.~\ref{eq:score} over the entire clip.  This asymmetry is
important context for interpreting the temporal-error
distribution below.

We measure the per-clip \emph{temporal error}
\begin{equation}
  \text{TE}_i \;=\; \bigl|\,f^\star_{\text{pred},i} - f^\star_{\text{gt},i}\,\bigr|
  \quad\text{(frames at $25$\,fps; $1$\,frame${=}40$\,ms)}.
  \label{eq:te}
\end{equation}

\paragraph*{Frame layout matters.}
Recall from Sec.~\ref{sec:samplers} that the contact-aware
sampler selects the $T{=}16$ frames \emph{asymmetrically}
around $f^\star_{\text{pred}}$: $9$ frames before, the contact
frame, and $6$ after, at stride $\tau{=}3$.  The effective
receptive field covers
$[\,f^\star_{\text{pred}} - 27,\;f^\star_{\text{pred}} + 18\,]$
frames.  The right metric for whether the sampler succeeds at
its job is therefore not whether $f^\star_{\text{pred}}$
matches the human-annotated frame exactly, but whether
$f^\star_{\text{gt}}$ falls inside this window so that the
classifier sees the contact event.

\paragraph*{Results.}
Table~\ref{tab:te-results} reports the per-clip comparison.

\begin{table}[!t]
\caption{Frame-selector evaluation on $20$ hand-annotated clips
($10$ Foul, $10$ Dive).  Frame indices are absolute
clip-frame indices.  TE is in frames at $25$\,fps.  ``In
window'' indicates whether the human-annotated frame lies
inside the model's $16$-frame window centred at
$f^\star_{\text{pred}}$ (i.e.\ inside
$[\,f^\star_{\text{pred}} - 27, f^\star_{\text{pred}} + 18\,]$).}
\label{tab:te-results}
\centering
\setlength{\tabcolsep}{4pt}
\renewcommand{\arraystretch}{1.05}
\begin{tabular}{llcccc}
\toprule
\textbf{Clip ID} & \textbf{Class} & \textbf{$f^\star_{\text{gt}}$} & \textbf{$f^\star_{\text{pred}}$} & \textbf{TE} & \textbf{In win.} \\
\midrule
Foul-1   & Foul & $73$ & $91$ & $18$ & \checkmark \\
Foul-2   & Foul & $69$ & $60$ & $9$  & \checkmark \\
Foul-3   & Foul & $66$ & $48$ & $18$ & \checkmark \\
Foul-4   & Foul & $76$ & $84$ & $\;\;8$  & \checkmark \\
Foul-5   & Foul & $66$ & $78$ & $12$ & \checkmark \\
Foul-6   & Foul & $75$ & $93$ & $18$ & \checkmark \\
Foul-7   & Foul & $46$ & $22$ & $24$ & --- \\
Foul-8   & Foul & $72$ & $90$ & $18$ & \checkmark \\
Foul-9   & Foul & $73$ & $91$ & $18$ & \checkmark \\
Foul-10  & Foul & $75$ & $62$ & $13$ & \checkmark \\
\midrule
Dive-1   & Dive & $15$ & $15$ & $\;\;0$  & \checkmark \\
Dive-2   & Dive & $73$ & $63$ & $10$ & \checkmark \\
Dive-3   & Dive & $77$ & $92$ & $15$ & \checkmark \\
Dive-4   & Dive & $78$ & $78$ & $\;\;0$  & \checkmark \\
Dive-5   & Dive & $70$ & $73$ & $\;\;3$  & \checkmark \\
Dive-6   & Dive & $74$ & $65$ & $\;\;9$  & \checkmark \\
Dive-7   & Dive & $76$ & $76$ & $\;\;0$  & \checkmark \\
Dive-8   & Dive & $\;\;9$  & $27$ & $18$ & \checkmark \\
Dive-9   & Dive & $38$ & $38$ & $\;\;0$  & \checkmark \\
Dive-10  & Dive & $62$ & $68$ & $\;\;6$  & \checkmark \\
\bottomrule
\end{tabular}
\end{table}

\begin{table}[!t]
\caption{Summary statistics across the $20$ annotated clips,
split by class.  Median is more representative than mean here
because the distribution is bimodal --- Dive clips concentrate
at low TE while Foul clips spread across the contact event.}
\label{tab:te-summary}
\centering
\setlength{\tabcolsep}{6pt}
\begin{tabular}{lccc}
\toprule
\textbf{Subset} & \textbf{Mean TE} & \textbf{Median TE} & \textbf{Max TE} \\
\midrule
All clips ($n{=}20$)   & $10.8$ & $11.0$ & $24$ \\
Foul clips ($n{=}10$)  & $15.6$ & $18.0$ & $24$ \\
Dive clips ($n{=}10$)  & $\;\,6.1$ & $\;\,4.5$ & $18$ \\
\bottomrule
\end{tabular}
\end{table}

\paragraph*{The headline result: window coverage is $\mathbf{19/20}$.}
In $19$ of $20$ clips the human-annotated contact frame lies
inside the $16$-frame window the contact-aware sampler selects.
This is the metric that matters for downstream classification:
whether or not $f^\star_{\text{pred}}$ exactly equals
$f^\star_{\text{gt}}$, the asymmetric $9{+}1{+}6$ frame layout
at stride $3$ reaches $27$ frames into the past and $18$ into
the future, which is wide enough to envelope the contact event
in almost every clip.  Only Foul-7 ($\text{TE}{=}24$) falls
outside the window.

\paragraph*{The Foul/Dive asymmetry.}
The mean and median TE are noticeably higher for Foul clips
(median $18$ frames) than for Dive clips (median $4.5$ frames).
This is consistent with the structural difference between the
two classes:
\begin{itemize}
  \item A Dive clip typically contains \emph{one} unambiguous
        salient event --- the onset of the simulated fall ---
        because by definition there is no real foul-causing
        contact~\cite{weiss2012kinematic}.  The smoothed combined score $\tilde{s}_f$ is
        nearly unimodal and both human and model converge on
        the same instant.  Six of the ten Dive clips have
        $\text{TE} \le 3$ frames; four are exact matches.
  \item A Foul clip typically contains \emph{multiple} contact
        events: the run-up jostling, shoulder-to-shoulder
        contact, the foul-causing contact itself, and the
        subsequent fall.  All of these produce peaks in
        $\tilde{s}_f$.  The human annotator semantically picks
        the foul-\emph{causing} contact; the algorithm picks the
        strongest signal peak, which may correspond to a
        different contact in the same event chain.  Hence the
        wider TE distribution on Foul clips.
\end{itemize}

This asymmetry is a known limitation of any heuristic
contact detector that lacks semantic understanding of intent.
It does not, however, prevent the classifier from succeeding:
as long as the foul-causing contact is inside the $16$-frame
window (which holds for $9$ of the $10$ Foul clips here), the
backbone sees enough of the event to make the binary decision.

\paragraph*{Failure case.}
The single out-of-window clip is Foul-7
($f^\star_{\text{gt}}{=}46$, $f^\star_{\text{pred}}{=}22$,
$\text{TE}{=}24$).  Inspection shows that an early run-up
collision between two players produces a sharper saliency peak
than the later foul-causing contact, attracting the $\arg\max$
even after smoothing and the approach-window guard.  This is
the second failure mode we identified during pipeline
development: \emph{double-event clips} where a non-foul
collision precedes the foul-causing one.  The first failure
mode --- YOLO detection drop-out during fast pans --- did not
occur in this $20$-clip sample.

\subsection{Adaptive Crop vs.\  Human Annotations}
\label{sec:iou-eval}

For the $20$ clips already used in the frame-selector
evaluation (Sec.~\ref{sec:te-eval}), three random people independently drew a square
axis-aligned ground-truth box around the offence region on the
human-annotated contact frame, at native $1280{\times}720$
resolution.  For each clip the element-wise mean of the three
boxes is taken as the consensus reference, and the model's
adaptive crop $\mathcal{C}_{\text{pred}}$ is compared against
it using intersection-over-union:
\begin{equation}
  \text{IoU} \;=\;
    \frac{|\mathcal{C}_{\text{pred}} \cap \mathcal{C}_{\text{gt}}|}
         {|\mathcal{C}_{\text{pred}} \cup \mathcal{C}_{\text{gt}}|}.
  \label{eq:iou}
\end{equation}
We adopt three operating bands: \textbf{High}
($\text{IoU}{\ge}0.7$, the crop is correctly anchored on the
offence); \textbf{Moderate} ($0.5{\le}\text{IoU}{<}0.7$, the
offence is captured but the box is offset); and \textbf{Poor}
($\text{IoU}{<}0.5$, the box misses a key actor).

\paragraph*{Inter-annotator agreement.}
Table~\ref{tab:iou-inter} reports inter-annotator IoU
statistics on the $20$-clip set under three increasingly
strict definitions.  Across all $20$ clips and $60$ pairwise
IoUs, mean inter-annotator agreement is
$0.965 \pm 0.062$.  This figure is partially inflated by $7$
clips on which all three annotators produced byte-identical
coordinates, consistent with the Pixlab tool snapping to
detected bounding boxes.  Restricted to the $13$ clips on
which at least one pair of annotators disagrees, mean
agreement is $0.946 \pm 0.071$ ($n{=}39$ pairs); restricted
further to the $7$ clips on which all three pairwise
annotations differ (the strictest independent-agreement
estimate), mean agreement is $0.934 \pm 0.065$ ($n{=}21$
pairs).  We treat $\approx 0.93$ as the realistic human ceiling
for this task.

\begin{table}[!t]
\caption{Inter-annotator IoU on the $20$-clip adaptive-crop
evaluation set, three subsets.  Pairs refers to all three
pairwise combinations per clip.  The third row is the
strictest estimate because it excludes clips where annotators
produced identical coordinates.}
\label{tab:iou-inter}
\centering
\setlength{\tabcolsep}{4pt}
\renewcommand{\arraystretch}{1.1}
\begin{tabular}{lccc}
\toprule
\textbf{Subset} & \textbf{Clips} & \textbf{Pairs} & \textbf{Mean$\pm\sigma$ IoU} \\
\midrule
All clips                                 & $20$ & $60$ & $0.965 \pm 0.062$ \\
Clips with any disagreement               & $13$ & $39$ & $0.946 \pm 0.071$ \\
All-pair disagreement (strict)            & $\;\;7$ & $21$ & $0.934 \pm 0.065$ \\
\bottomrule
\end{tabular}
\end{table}

\begin{figure*}[!t]
\centering
\includegraphics[width=0.95\textwidth]{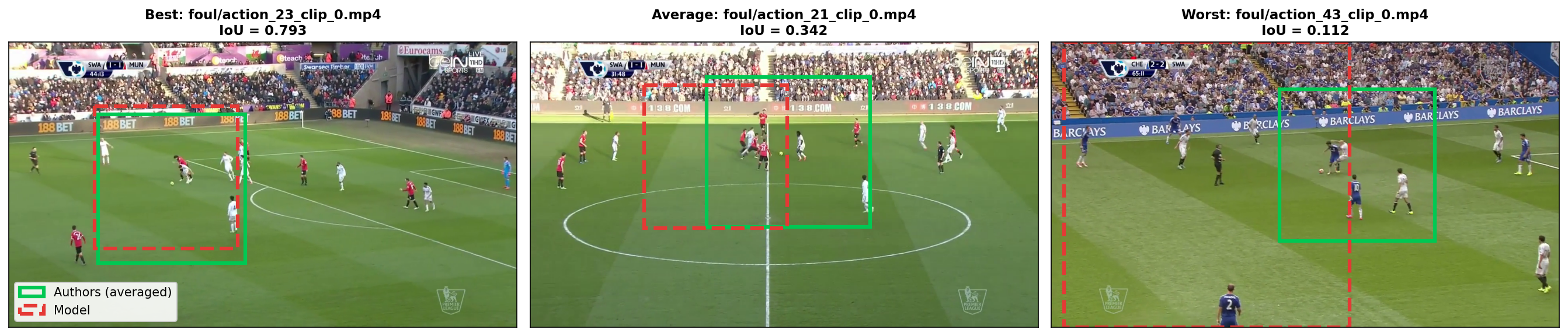}
\caption{Adaptive-crop qualitative cases at the human-annotated
contact frame.  Green solid: averaged author box (element-wise
mean of three independent annotations); red dashed: predicted
crop from the \emph{adaptive crop} module.}
\label{fig:iou-cases}
\end{figure*}

\paragraph*{Model vs averaged annotator box.}
Table~\ref{tab:iou-results} reports the model's adaptive crop
versus the averaged random people box across $20$ clips, and
Fig.~\ref{fig:iou-cases} shows representative best, average,
and worst cases.  Mean model-vs-authors IoU is
$0.369 \pm 0.207$ over $n{=}20$ clips, with peak IoU $0.793$
and minimum $0.112$.  The distribution is bimodal: $3$ clips
reach High alignment ($\text{IoU}{\ge}0.7$), $2$ Moderate, and
$15$ Poor.  The gap between the model and the human ceiling
($\approx 0.93$) is substantial, and we discuss its dominant
cause below.

\begin{table}[!t]
\caption{Adaptive-crop evaluation summary.  Model crop vs
averaged random people box across $20$ clips.  Per-annotator IoUs are
nearly identical (mean $0.369$, $0.366$, $0.365$ for the three
annotators respectively), consistent with the high
inter-annotator agreement reported in
Table~\ref{tab:iou-inter}.}
\label{tab:iou-results}
\centering
\setlength{\tabcolsep}{4pt}
\renewcommand{\arraystretch}{1.1}
\begin{tabular}{lc}
\toprule
\textbf{Quantity} & \textbf{Value} \\
\midrule
Mean model-vs-random people IoU& $0.369 \pm 0.207$ \\
Median model-vs-random people IoU& $0.312$ \\
Maximum model-vs-random people IoU& $0.793$ \\
Minimum model-vs-random people IoU& $0.112$ \\
\midrule
\multicolumn{2}{l}{\emph{Alignment band counts ($n{=}20$):}} \\
\quad High      ($\text{IoU}{\ge}0.7$)        & $\;3$ \\
\quad Moderate  ($0.5{\le}\text{IoU}{<}0.7$)  & $\;2$ \\
\quad Poor      ($\text{IoU}{<}0.5$)          & $15$ \\
\bottomrule
\end{tabular}
\end{table}

\paragraph*{Failure mode.}
The dominant failure mode of the adaptive crop is YOLO
detection drop-out~\cite{vats2021football,vandeghen2022kalman}.  The pipeline composes ball, holder,
opponent, fall, and leg-overlap anchors into a priority
centroid (Sec.~\ref{sec:adaptive-crop}); when any of the three
primary anchors (ball, holder, opponent) is missing on the
contact frame, the centroid is pulled by whichever anchors
remain, producing a lateral shift that clips part of the
offence.  Inspection of the $15$ Poor-band clips shows ball
drop-out (small object, fast pans) as the dominant cause.
This is the same root cause as the single out-of-window
frame-selector failure (Sec.~\ref{sec:te-eval}): both stem
from YOLO detection reliability on small / fast-moving
targets, and replacing YOLO with a more robust
ball-and-player tracker is the most direct target for future
improvement (Sec.~\ref{sec:limitations}).

\paragraph*{What this tells us.}
The result is honest rather than flattering.  At the current pipeline stage, 
and with the current YOLO detector, the adaptive crop attains High alignment 
with author annotations only when all three
primary anchors are detected reliably; on the majority of
clips it falls in the Poor band.  Two things qualify this.
First, the binary-classification accuracy result
(Table~\ref{tab:f1-test}: $86\%$ test, $0.860$ macro-F$1$) is
attained \emph{despite} the imperfect crop, indicating that
the classifier tolerates moderate spatial misalignment as long
as the offence is partially in-frame --- which is consistent
with the contact-aware sampler's window-coverage result of
$19/20$.  Second, the failure mode is single-sourced (YOLO
anchor drop-out), which makes it a concrete target for a
future-work step rather than a fundamental limitation of the
contact-aware design.

\subsection{Per-Class Behaviour}
\label{sec:perclass}

\begin{figure}[!t]
\centering
\includegraphics[width=0.6\columnwidth]{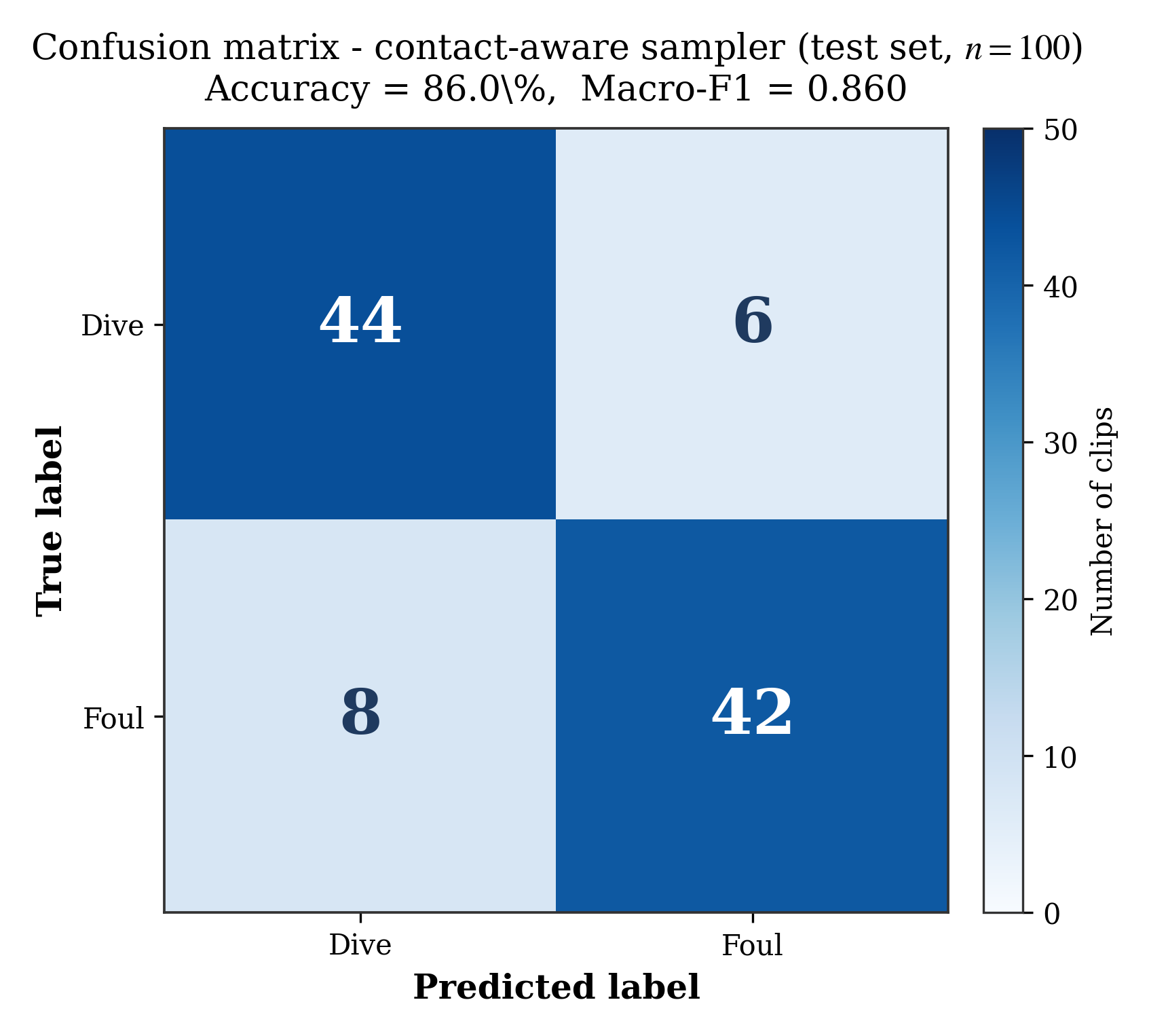}
\caption{Test-set confusion matrix of the contact-aware
sampler (seed $42$, $100$ held-out clips).
Macro-F$1{=}0.860$.}
\label{fig:confmat-test}
\end{figure}

Figure~\ref{fig:confmat-test} shows the test-set confusion
matrix of the contact-aware sampler.  Per-class performance is
near-symmetric on the held-out split: Dive precision/recall
$=0.846/0.880$, Foul precision/recall $=0.875/0.840$,
macro-F$1$ $=0.860$.  Looking across all four samplers in
Table~\ref{tab:f1-test}, only the contact-aware sampler
achieves balanced Foul precision and recall (both
${\geq}0.84$); the three content-blind baselines exhibit
asymmetric per-class behaviour --- e.g.\ Random reaches
$R_{\text{Foul}}{=}0.88$ but only $R_{\text{Dive}}{=}0.62$,
indicating it over-predicts Foul on ambiguous clips.  This
asymmetric Foul-vs-Dive bias is the characteristic failure
mode of content-blind processing on contact-driven actions,
and the same effect is visible in the earlier optical-flow
baseline of Al-Theiabat and
Aljarrah~\cite{altheiabat2018dive}, where all four
mis-classifications were Foul predicted as Dive.  The
contact-aware sampler corrects this asymmetry.

\subsection{Convergence and Stability}

\begin{figure}[!t]
\centering
\includegraphics[width=\columnwidth]{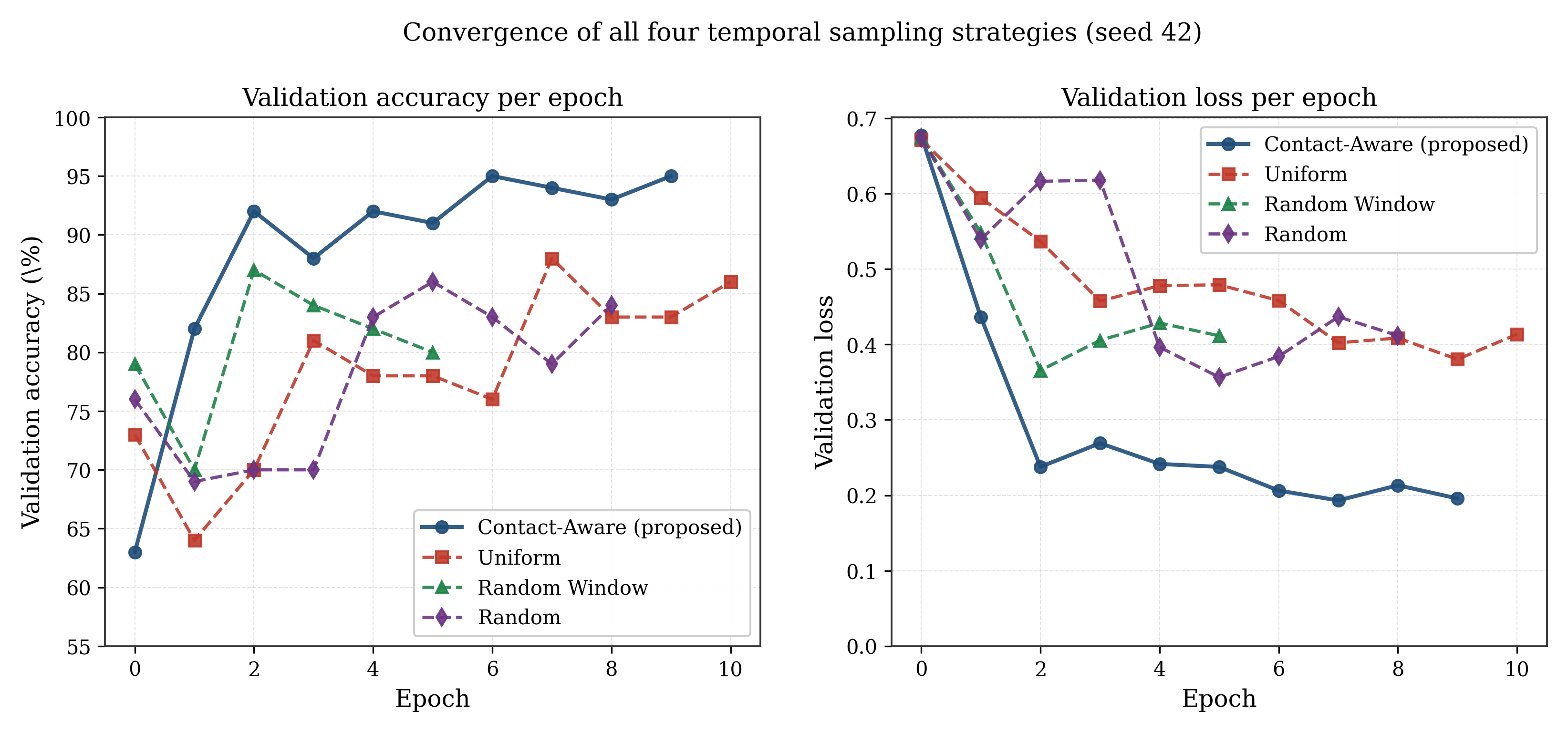}
\caption{Validation accuracy (left) and loss (right) for all
four samplers, seed $42$.}
\label{fig:curves}
\end{figure}

Figure~\ref{fig:curves} shows the validation accuracy and loss
curves under seed $42$, used for model selection.  The
contact-aware sampler reaches $92\%$ by epoch $2$ and $95\%$
at epoch $6$ and stabilises without oscillation; its
validation loss decreases monotonically from $0.677$ at epoch
$0$ to $0.196$ at epoch $9$ (best $0.206$ at epoch $6$).  The
three baselines converge more slowly and are more volatile.

\subsection{Qualitative Frame Selection vs.\ Uniform}

\begin{figure*}[!t]
\centering
\includegraphics[width=0.95\textwidth]{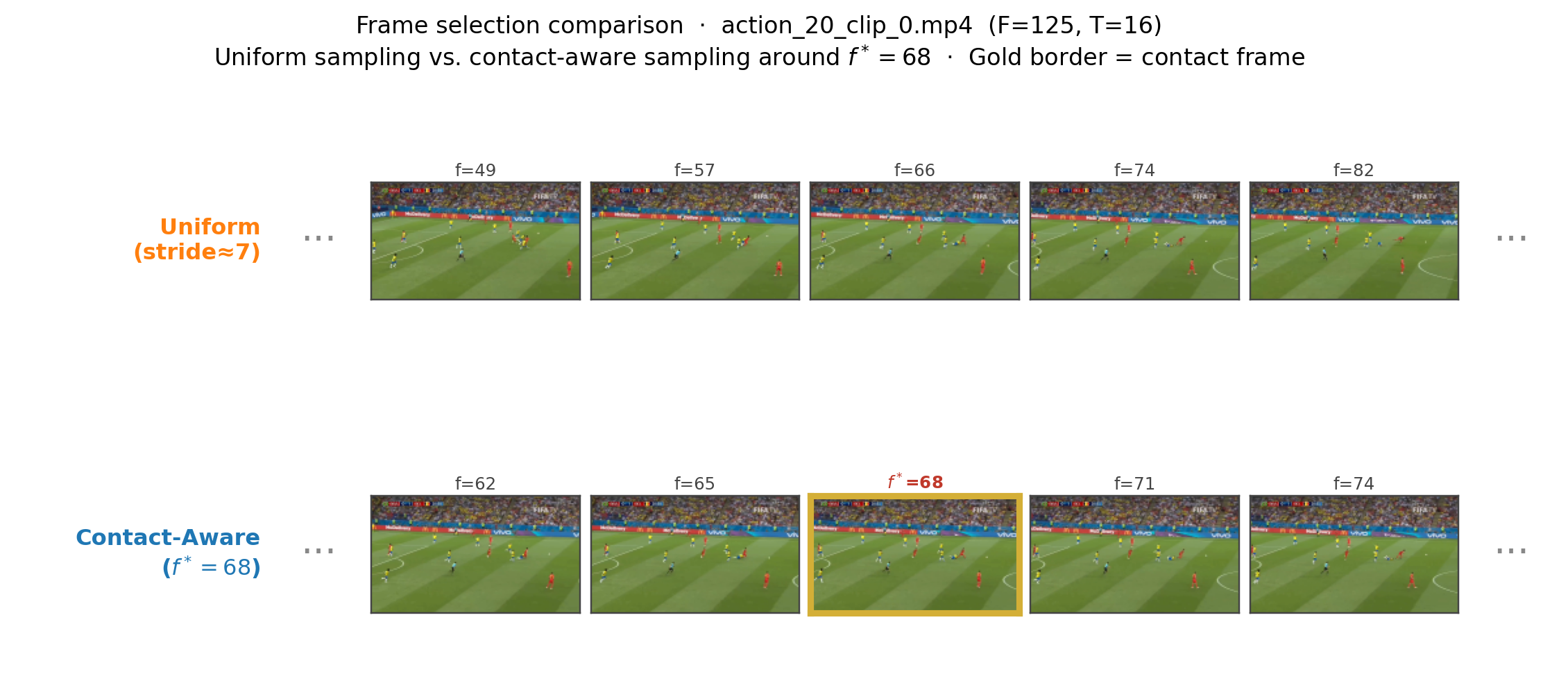}
\caption{Frame indices selected by uniform (top) and
contact-aware (bottom) on the same Dive clip
($F{\approx}125$).}
\label{fig:frame-selection}
\end{figure*}

Figure~\ref{fig:frame-selection} contrasts the frames selected
by uniform sampling and by the contact-aware sampler on a
representative Dive clip.  Uniform sampling spreads its
$T{=}16$ frames across $F{\approx}125$ at stride ${\approx}8$;
the contact-aware sampler concentrates them inside an
$\approx 1.8$\,s window centred on $f^\star{=}66$, so every
selected frame lies inside the discriminative segment.

\subsection{Reproducibility and Seed Stability}
\label{sec:repro}

The primary experimental protocol (Sec.~\ref{sec:experiments})
fixes the random seed at $42$.  To verify that the seed-$42$
ranking is not a consequence of a fortunate initialisation, we
re-train the proposed contact-aware sampler and the strongest
baseline (uniform) under two additional seeds $\{7, 123\}$,
keeping every other aspect of the training recipe identical.
Random and random-window are not re-trained because the
purpose of this check is to validate the proposal against the
strongest baseline, not to re-rank the weaker baselines that
already lose under the primary protocol.

Table~\ref{tab:repro} reports the best-epoch validation
accuracy under each seed.  Three observations support the
seed-$42$ conclusion:
\begin{itemize}
  \item \textbf{Within-sampler variance is small relative to
        the between-sampler gap.}  The standard deviation
        across the three seeds is $\sigma_{\text{CA}}{=}2.1$\,pp
        and $\sigma_{\text{Uniform}}{=}2.9$\,pp, both well below
        the $8.0$\,pp mean gap.
  \item \textbf{The ranking is preserved on every seed.}  The
        per-seed gap (CA $-$ Uniform) is $\{7, 8, 9\}$\,pp,
        all of the same sign; the worst contact-aware seed
        ($92\%$, seed $123$) still exceeds the best uniform
        seed ($88\%$, seeds $42$ and $7$) by $4$\,pp.
  \item \textbf{The improvement is statistically supported.}
        A paired $t$-test on the three per-seed gaps yields
        $t{=}13.9$ ($df{=}2$, $p{=}0.005$), significant at
        $\alpha{=}0.01$.  We treat this as corroborative
        rather than definitive given the small seed count.
\end{itemize}

These runs are reported strictly to validate reproducibility
of the seed-$42$ result; they are not used to compute the
per-class metrics, confusion matrix, or qualitative analyses
elsewhere in the paper, which remain seed-$42$-only.

\begin{table}[!t]
\caption{Seed-stability check for the proposed sampler and its
strongest baseline: best validation accuracy under three seeds.
The primary experimental result (Table~\ref{tab:results}) uses
seed $42$; the additional seeds $\{7, 123\}$ verify that the
seed-$42$ ranking is robust.  $^{\dagger}$Training interrupted
at epoch $8$ before early stopping triggered; the value is the
best validation accuracy at the saved checkpoint (epoch $7$).}
\label{tab:repro}
\centering
\setlength{\tabcolsep}{4pt}
\renewcommand{\arraystretch}{1.1}
\begin{tabular}{lcccc}
\toprule
\textbf{Sampler} & \textbf{Seed 42} & \textbf{Seed 7} & \textbf{Seed 123} & \textbf{Mean$\pm\sigma$} \\
\midrule
Cont.-Aware \emph{(prop.)} & $95.0$ & $96.0^{\dagger}$ & $92.0$ & $\mathbf{94.3 \pm 2.1}$ \\
Uniform                    & $88.0$ & $88.0$           & $83.0$ & $86.3 \pm 2.9$ \\
\midrule
\multicolumn{4}{l}{\emph{Per-seed gap (CA $-$ Uniform):}} & $\{7, 8, 9\}$ \\
\multicolumn{4}{l}{\emph{Paired $t$-test:}}                & $t{=}13.9$, $p{=}0.005$ \\
\bottomrule
\end{tabular}
\end{table}

\section{Discussion}
\label{sec:discussion}

\subsection{Why Contact-Awareness Matters}
The foul/dive distinction is a fine-grained action whose
discriminative information lies almost entirely in a few
hundred milliseconds around the instant of contact~\cite{weiss2012kinematic}.  Uniform
sampling, while simple, has only a $T/F \approx 13\%$ chance
of placing any one of its $T{=}16$ frames in the most
informative half-second of a $5$\,s clip.  By anchoring the
sampling window on a YOLO-derived contact estimate, the
contact-aware sampler systematically increases the probability
that the model sees the moment that decides the label.

The test-set evaluation reinforces this argument empirically.
Whereas all four samplers see a drop from validation to test
accuracy --- a normal consequence of using val for model
selection --- the magnitude of that drop differs.  The
contact-aware sampler loses $9$\,pp (val $95\%$ $\to$ test
$86\%$), random-window loses $8$\,pp, but uniform loses
$14$\,pp (val $88\%$ $\to$ test $74\%$).  Content-driven
sampling appears to over-fit to the validation split less
than content-blind sampling, plausibly because the
contact-anchored window is determined by the YOLO-driven
signal rather than by training-data idiosyncrasies, leaving
fewer degrees of freedom for the classifier to exploit.

The qualitative per-component evaluations
(Sec.~\ref{sec:te-eval}--\ref{sec:iou-eval}) make this
concrete.  On the $20$-clip annotated subset the asymmetric
$16$-frame window selected by the sampler covers the
human-annotated contact frame in $19$ of $20$ cases,
indicating that the sampler reliably places the
discriminative event inside the frames the classifier sees.
The exact-frame agreement is tighter on Dive clips (median
$\text{TE}{=}4.5$ frames, with $4/10$ exact matches) than on
Foul clips (median $\text{TE}{=}18$ frames), because Foul
clips contain multiple contact events --- run-up jostling,
shoulder-to-shoulder contact, foul-causing contact --- and
the algorithm picks the strongest signal peak whereas the
human picks the foul-\emph{causing} peak semantically.  This
Foul/Dive asymmetry is consistent with the nature of the two
classes and is not a malfunction.  On the $20$-clip
three-annotator adaptive-crop comparison the model attains
mean $\text{IoU}{=}0.369 \pm 0.207$ against the averaged
author box, against an inter-annotator ceiling of
$\approx 0.93$.  Three clips reach the High alignment band
and $15$ fall in the Poor band; inspection traces the
Poor-band behaviour to YOLO anchor drop-out --- the same
root cause as the single out-of-window frame-selector
failure --- meaning the adaptive crop is bounded by the
underlying detector rather than by the priority-centroid
design itself.
Figure~\ref{fig:signal-trace} provides the mechanistic
counterpart: the four detection-derived signals exhibit
qualitatively different temporal profiles on Foul vs.\ Dive
clips, which is why the smoothed combined score $\tilde{s}$
produces a cleanly contact-aligned $f^\star$ on most clips.

\subsection{Why Single-View Matters}
The contact-aware sampler can be read as a single-view
substitute for the contact disambiguation that multi-view
pipelines obtain for free from a close-up replay angle.  In a
multi-view setup such as SoccerNet-MVFouls, a close-up replay
typically resolves the contact moment by virtue of being
aimed at the offence.  In our single-view setting no such
auxiliary view is available, so the model itself must answer
\emph{when} and \emph{where}.  The sampler externalises both
questions to a lightweight YOLO detector: \emph{when}, by
maximising $\tilde{s}_f$; \emph{where}, by anchoring the
priority crop on the detected ball, holder, and falling
player.

\subsection{Cost, Memory, and Trade-offs}
The contact-aware sampler is approximately $40\%$ slower per
training step than uniform on T$4$ hardware because it runs
YOLO on every frame.  Its dominant memory cost is holding the
entire decoded clip in CPU RAM ($\approx 1.5$\,GB at
$1280{\times}720$, $25$\,fps, $5$\,s), which forces
single-process data loading on T$4$.  At inference time the sampler runs once per clip, so the amortised cost is small.

\subsection{Limitations}
\label{sec:limitations}
First, our YOLO detector inherits its failure modes
(occlusion, motion blur, low-contrast kit colours)~\cite{vats2021football,vandeghen2022kalman}, and these
account for the worst-case temporal-error and IoU outliers in
Sec.~\ref{sec:te-eval}--\ref{sec:iou-eval}; the same clip
(Foul-7) is the worst case in both evaluations, and the shared
root cause is YOLO anchor drop-out.  Second, the dataset is
modest ($600$ clips) from professional broadcast footage;
generalisation to amateur or alternative-competition footage
is not evaluated.  Third, the per-component evaluations remain small-scale
. The adaptive-crop evaluation uses $20$
clips and three independent student annotators, all of whom
are co-authors of this paper; annotators external to the
author team, together with a larger annotated subset, would
strengthen the certificates against shared methodological
biases.  The current $20$-clip set reaches the High alignment
band only on $3$ clips; a larger subset would more precisely
characterise how often the pipeline succeeds in absolute
terms versus how often it is bottlenecked by YOLO detection
reliability.  Fourth, the seed-stability check
(Sec.~\ref{sec:repro}) is restricted to the proposed sampler
and its strongest baseline at $n{=}3$ seeds; a fully matched
four-sampler $\times$ multi-seed factorial would tighten the
ranking among the weaker baselines (though it would not
affect the central proposal-vs-uniform claim, which is
already validated under three seeds and a paired test).
Fifth, we cannot directly compare our test-set numbers to the
$84\%$ $10$-fold cross-validation accuracy reported by
Al-Theiabat and Aljarrah~\cite{altheiabat2018dive}, because
their $25$-clip dataset is not publicly released and their
pipeline requires per-clip manual selection of the falling
player.  We can only compare protocols and per-class
behaviour, as discussed in Sec.~\ref{sec:perclass}.  Sixth,
we evaluate only the binary case; a complete refereeing
assistant would need severity and intent labels~\cite{ifab2023laws}.  These are
explicit limitations, not hidden ones; they shape the
realistic claims this paper can make.

\section{Conclusion and Future Work}
\label{sec:conclusion}

We presented a contact-aware temporal-sampling and
spatial-cropping front-end for VideoMAE fine-tuning on the
single-view foul-vs-dive recognition task.  The contribution
is engineering and presentation, not architectural: a
documented $600$-clip single-view dataset (approximately
$24\times$ larger than the only previous single-view
Foul/Dive corpus~\cite{altheiabat2018dive}), a reproducible
end-to-end automated pipeline, and a qualitative
per-component evaluation that compares predicted contact
frames and predicted crops against human annotation.  On the
held-out test split the contact-aware sampler attains $86\%$
accuracy and macro-F$1{=}0.860$ (vs.\ $74\%$ and $0.740$ for
uniform, a $12$\,pp gap that is larger than the
corresponding $7$\,pp validation-set gap and that holds across
three random seeds, paired $t{=}13.9$, $p{=}0.005$).  On the
$20$-clip annotated subset the model's asymmetric $16$-frame
window covers the human-annotated contact frame in $19$ of
$20$ cases, with Dive clips showing tight agreement and Foul
clips a wider spread caused by the multiple contact events
typical of a foul sequence.  On the $20$-clip three-annotator
adaptive-crop comparison the model attains mean
$\text{IoU}{=}0.369 \pm 0.207$ against the averaged author
box ($3/20$ High, $2/20$ Moderate, $15/20$ Poor), with the
Poor-band behaviour traceable to YOLO anchor drop-out ---
the same root cause as the single out-of-window
frame-selector failure --- making the failure mode
diagnosable and a concrete target for future improvement.
Within the limitations of a focused small-scale study, the
front-end blocks behave as designed and their failure modes
are diagnosable.

Natural extensions include: extending the multi-seed protocol
to all four samplers and a larger seed count; a larger
annotated subset with annotators external to the author team;
replacing the YOLO detector with a more robust
ball-and-player tracker~\cite{vandeghen2022kalman} to address the anchor drop-out
failure mode identified above; a learned aggregator to
replace the hand-tuned signal weights; football-domain
self-supervised pre-training with VideoMAE
V2~\cite{wang2023videomaev2}; multi-class extension
to foul severity and intent~\cite{ifab2023laws}; and cross-domain evaluation on
amateur footage.

\begin{acks}
The authors thank the \emph{Stop Diving} initiative for
maintaining the publicly accessible catalogue of documented
dive incidents that supported the curation of the dive subset
of the dataset~\cite{stopdiving}.
\end{acks}

\bibliographystyle{ACM-Reference-Format}
\bibliography{References}

\end{document}